\documentclass[final,5p,times,twocolumn]{elsarticle}
\usepackage{amssymb}
\usepackage{lineno,hyperref}
\usepackage[np, autolanguage]{numprint}
\modulolinenumbers[5]

\usepackage{tikz}
\newcommand*\circled[1]{\tikz[baseline=(char.base)]{
            \node[shape=circle,draw,inner sep=1pt,font=\sffamily\footnotesize] (char) {\textbf{#1}};}}
            
\usepackage{amsmath,amssymb,amsfonts}
\usepackage{booktabs}
\usepackage{multirow}
\usepackage{tabularx}
\usepackage{tabulary}
\usepackage{nicefrac}
\usepackage{soul}
\usepackage{comment}
\usepackage{xcolor}

\usepackage{amsthm}

\usepackage{lineno}

\usepackage{color}
\usepackage{pdfcomment}
\definecolor{burntorange}{rgb}{0.8, 0.33, 0.0}
\usepackage{todonotes}

\journal{Journal of Systems and Software}

\begin{document}

\begin{frontmatter}

%% Title, authors and addresses

%% use the tnoteref command within \title for footnotes;
%% use the tnotetext command for theassociated footnote;
%% use the fnref command within \author or \affiliation for footnotes;
%% use the fntext command for theassociated footnote;
%% use the corref command within \author for corresponding author footnotes;
%% use the cortext command for theassociated footnote;
%% use the ead command for the email address,
%% and the form \ead[url] for the home page:
%% \title{Title\tnoteref{label1}}
%% \tnotetext[label1]{}
%% \author{Name\corref{cor1}\fnref{label2}}
%% \ead{email address}
%% \ead[url]{home page}
%% \fntext[label2]{}
%% \cortext[cor1]{}
%% \affiliation{organization={},
%%            addressline={}, 
%%            city={},
%%            postcode={}, 
%%            state={},
%%            country={}}
%% \fntext[label3]{}

\title{Enhancing the Analysis of Software Failures\\ in Cloud Computing Systems with Deep Learning}

%% use optional labels to link authors explicitly to addresses:
%% \author[label1,label2]{}
%% \affiliation[label1]{organization={},
%%             addressline={},
%%             city={},
%%             postcode={},
%%             state={},
%%             country={}}
%%
%% \affiliation[label2]{organization={},
%%             addressline={},
%%             city={},
%%             postcode={},
%%             state={},
%%             country={}}

\author{Domenico Cotroneo}
\ead{cotroneo@unina.it}
\author{Luigi De Simone}
\ead{luigi.desimone@unina.it}
\author{Pietro Liguori}
\ead{pietro.liguori@unina.it}
\author{Roberto Natella}
\ead{roberto.natella@unina.it}

\address{Università degli Studi di Napoli Federico II}
\address{Naples, Italy}
%\affiliation{organization={a},
%            addressline={a}, 
%            city={Naples},
%            postcode={a}, 
%            state={Italy},
%            country={a}}

\begin{abstract}
Identifying the failure modes of cloud computing systems is a difficult and time-consuming task, due to the growing complexity of such systems, and the large volume and noisiness of failure data.
This paper presents a novel approach for analyzing failure data from cloud systems, in order to relieve human analysts from manually fine-tuning the data for feature engineering. The approach leverages Deep Embedded Clustering (DEC), a family of unsupervised clustering algorithms based on deep learning, which uses an autoencoder to optimize data dimensionality and inter-cluster variance. 
We applied the approach in the context of the OpenStack cloud computing platform, both on the raw failure data and in combination with an anomaly detection pre-processing algorithm.
%We evaluated the approach on failure data from the OpenStack cloud computing platform. 
The results show that the performance of the proposed approach, in terms of purity of clusters, is comparable to, or in some cases even better than manually fine-tuned clustering, thus avoiding the need for deep domain knowledge and reducing the effort to perform the analysis. In all cases, the proposed approach provides better performance than unsupervised clustering when no feature engineering is applied to the data. Moreover, the distribution of failure modes from the proposed approach is closer to the actual frequency of the failure modes.
\end{abstract}

%%Graphical abstract
%\begin{graphicalabstract}
%\includegraphics{grabs}
%\end{graphicalabstract}

%%Research highlights
%\begin{highlights}
%\item Research highlight 1
%\item Research highlight 2
%\end{highlights}

\begin{keyword}
%% keywords here, in the form: keyword \sep keyword
Failure Mode Analysis \sep Software Failures \sep Fault Injection \sep Cloud Computing \sep Deep Learning \sep OpenStack 
%% PACS codes here, in the form: \PACS code \sep code

%% MSC codes here, in the form: \MSC code \sep code
%% or \MSC[2008] code \sep code (2000 is the default)

\end{keyword}

\end{frontmatter}

%% \linenumbers

%% main text
\section{Introduction}
\label{sec:introduction}
Nowadays, cloud computing fuels critical services for our life, such as telecom, healthcare, transportation, and other domains where high reliability is mandatory. However, cloud software systems can fail in unpredictable ways, due to cascading propagation of faults across their components \cite{garraghan2018emergent,hole2019software}. 
To prevent service outages, cloud system designers need to know in advance how their software system behaves under failure (\emph{failure modes}) before deploying it in operation. Knowledge of failure and repair characteristics is valuable for designers in order to plan failure management solutions \cite{vishwanath2010characterizing,li2006job,6754595}.

%In order to create solutions for improving the dependability of cloud computing systems, it is imperative to analyze data from real-world sources \cite{6754595}. For example, the state-of-the-art often relies on statistical properties and accurate modeling of failure and repair characteristics \cite{vishwanath2010characterizing,li2006job}.

To get data about software failures, \emph{fault injection} is typically adopted \cite{hsueh1997fault,arlat1990fault}, i.e., the deliberate insertion of \textit{faults} (such as resource exhaustion, software bugs, connection loss, etc.) into a software system in a controlled experiment, in order to trigger failures. 
These experiments produce a large amount of failure data, in terms of hundreds of thousands of events and execution traces. From this large amount of data, \emph{failure modes analysis} aims to identify which are the recurring failure modes and their relative frequency.
Such analysis guides the human designer towards prioritizing the development of failure management mechanisms for the most frequent and severe failure modes. 

Unfortunately, failure mode analysis is a difficult and time-consuming task, due to the size and complexity of failure data. 
Moreover, failure mode analysis is hindered by the non-deterministic behavior of cloud systems, which causes random variations in the timing and the ordering of events in the system, thus introducing noise in the failure data \cite{zhao2010fault}. Therefore, failure mode analysis techniques must be robust to noise in the failure data. 
The adoption of unsupervised machine learning techniques, such as clustering and anomaly detection, comes to the rescue but still faces some limitations. These techniques require the preliminary selection and transformation features (\emph{feature engineering}) \cite{mousavi2019unsupervised,zhang2020unsupervised,Xu2021FeatureSA}, to make the failure data more amenable for analysis. This effort requires deep domain knowledge and represents a significant up-front cost.

In this work, we propose a novel approach for efficiently identifying recurrent failure modes from failure data. The approach leverages deep learning for unsupervised machine learning, to overcome the challenges of noise and complexity of the feature space. 
Our approach saves the manual efforts spent on feature engineering, by using an autoencoder to automatically transform the raw failure data into a compact set of features. The approach transforms the data by jointly optimizing for the reconstruction error (i.e., the transformed features are still representative of the sample) and inter-cluster variance (i.e., to make it easier to identify groups of similar failures).

We evaluated the proposed approach on a dataset of thousands of failures from OpenStack \cite{OpenStack}, a popular platform used in several private and public cloud computing systems, and the basis of over 30 commercial products \cite{OpenStackProducts, OpenStackUsers}. As an additional contribution to this work, we publicly released this dataset for the research community. We compare the proposed approach to a manually fine-tuned clustering technique. The results demonstrate that the proposed approach can identify clusters with accuracy similar, or in some cases, even superior, to the fine-tuned clustering, with a low computational cost.

The paper is structured as follows: Section~\ref{sec:related} discuss the related work; Section~\ref{sec:background} provides details on the background; Section~\ref{sec:approach} presents the proposed approach; Section~\ref{sec:experiments} evaluates the approach; Section~\ref{sec:conclusion} concludes the paper.

\section{Related Work}
\label{sec:related}
\noindent
\textbf{Uncertainty in fault injection experiments.} Uncertainty is a key aspect in fault injection experimentation since the behavior of a complex system depends on many factors that are difficult or impossible to control. This problem is exacerbated when the fault-injection is used in cloud computing, where the human analyst has to deal with the non-deterministic nature of such systems.
State-of-the-art provides several works that addressed this problem by applying solutions based on statistical techniques.
Several studies leveraged the statistical models to model the probability of failures during hardware fault-injection experiments \cite{arlat2011collecting, skarin2010goofi, palazzi2019tale}.
Arlat \textit{et al.} \cite{arlat1993fault} proposed a solution that brings together the coverage evaluation of the fault coverage and the occurrence of the faults to estimate the dependability of the complex fault-tolerant systems.
By estimating the probabilities of the failure modes of the system, Voas \textit{et al.} \cite{voas1997reducing} presented a solution to reduce the uncertainty of whether different software faults impact the behavior of the system.
To assess the quality of the measurements in terms of uncertainty, repeatability, resolution, and intrusiveness, Bondavalli \textit{et al.} \cite{bondavalli2007foundations, bondavalli2010new} applied the principles of \emph{measurement theory}. 
In AMBER project \cite{wolter2012resilience}, the authors used data mining to identify the factors (i.e., workloads, the fault types, etc.) with the highest impact on the performance and availability of the target system.  
Loki tool \cite{DBLP:journals/tpds/ChandraLJCS04} uses an off-line clock synchronization algorithm to collect traces of events exchanged among the nodes, and performs a post-experiment analysis to identify if the injections hit the desired state, and repeats the experiments only when needed.
Gulenko \textit{et el.} \cite{gulenko2018detecting} introduced an anomaly detection approach that leverages an online clustering method to define the normal behavior of monitored components. Wu \textit{et al.} \cite{wu2020performance} proposed a method that applies a dependency graph and an autoencoder to identify the causes of the performance degradation in the microservices of the cloud. Both previous works evaluated the proposed solutions by injecting performance anomalies in the cloud computing system.
%The Loki tool \cite{DBLP:journals/tpds/ChandraLJCS04} addressed the problem of injecting faults in controlled global states of distributed systems since it is difficult due to the lack of a global clock and communication delays (e.g., between a central controller and a local injector). The tool performs a post-experiment analysis of event traces collected from nodes, using an off-line clock synchronization algorithm, to identify whether injections hit the desired state, and repeats the experiments only when needed.
%Hardware fault injection has been sampling the space of fault injections (i.e., CPU instructions and data words to be injected with \emph{bit-flips}), and applying statistical modeling for the probability of failures (e.g., to obtain confidence intervals) \cite{arlat2011collecting, skarin2010goofi, palazzi2019tale}. 
%This approach has been generalized in the AMBER project \cite{wolter2012resilience}, which used data mining techniques to analyze large sets of experiments, to identify the factors (e.g., the type of injected faults, the workload, the configuration of the target system, etc.) with the highest impact on performance and availability of the system. 
%These considerations are exacerbated when the fault-injection is applied in the distributed systems, where the non-deterministic behavior of these systems introduces additional uncertainty in experiments.
\\
All these studies are based on the assumption that failures can be accurately and automatically identified. We consider our work complementary to them since it provides novel techniques for identifying the failure modes of the target system.

\vspace{3pt}
\noindent
\textbf{Failure mode clustering.} 
The use of clustering to automatically discover and analyze failure modes is a topic widely addressed by previous research. 
Arunajadai \textit{et al.} \cite{arunajadai2004failure} described a clustering-based method for grouping failure modes in electromechanical consumer products. The approach groups failure modes based on their occurrence, to determine whether a failure should be considered by itself or whether it tends to accompany other kinds of failures. Then, the analyst can prioritize critical failure modes. The approach uses a hierarchical clustering algorithm with the complete linkage method. 
Chang \textit{et al.} \cite{chang2015clustering} combines clustering with risk management, by grouping failure modes that have similar risk levels concerning three factors (severity, occurrence, detection), and visualizes them to ease multi-criteria decision making. Their approach clusters and visualizes failures as a tree structure that is easy to understand. It is evaluated in the context of farming applications.  
Duan \textit{et al.} \cite{duan2019new} analyze evaluations of failure modes in natural language by FMEA experts, using fuzzy sets to extract features, and the \emph{k-means} algorithm to cluster the failure modes. 
Xu \textit{et al.} \cite{xu2020data} proposed a method to construct the component-failure mode (CF) matrix automatically, by mining unstructured texts using the Apriori algorithm and the semantic dictionary WordNet to build a standard set of failure modes. As in the work by Arunajadai et al. \cite{arunajadai2004failure}, the matrix is used for grouping the failure modes using clustering algorithms, such as the \emph{K-means}. 
Rahimi \textit{et al.} \cite{rahimi2019clustering} analyzed a large dataset of truck crash data, based on police reports about the driver, vehicle, crash, and citation information. They address the problem of high-dimensionality spaces, by adopting block clustering to investigate heterogeneity in the crash dataset. This approach considers two sets (observations and variables) simultaneously and organizes the data into homogeneous blocks.
Liu \textit{et al.} contributed with several studies on the failure mode and effects analysis \cite{huang2020failure}. They improved failure mode analysis using two-dimensional uncertain linguistic variables and alternative queuing \cite{liu2018failure} and proposed a novel approach combining HULZNs and DBSCAN algorithms to assess and cluster the risk of failure modes \cite{liu2020new}. They evaluated the feasibility of the proposed approaches in real use-case scenarios, showing the ability to classify failure modes in complex and uncertain conditions.

Different from these solutions, our approach is tailored for the domain of cloud system failures, where the data consist of symbolic sequences, which are obtained from events recorded through distributed tracing technology. Our approach leverages the deep neural networks, to automatically cluster the failure modes without manual effort for feature engineering. Moreover, we also investigate clustering in combination with anomaly detection for cloud systems.

\section{Background}
\label{sec:background}
This section provides information on cloud computing systems, with emphasis on OpenStack, on failure mode analysis, and on the open issues that are addressed in this work. 

\begin{figure*}[ht]
    \centering
    \includegraphics[width=1.75\columnwidth]{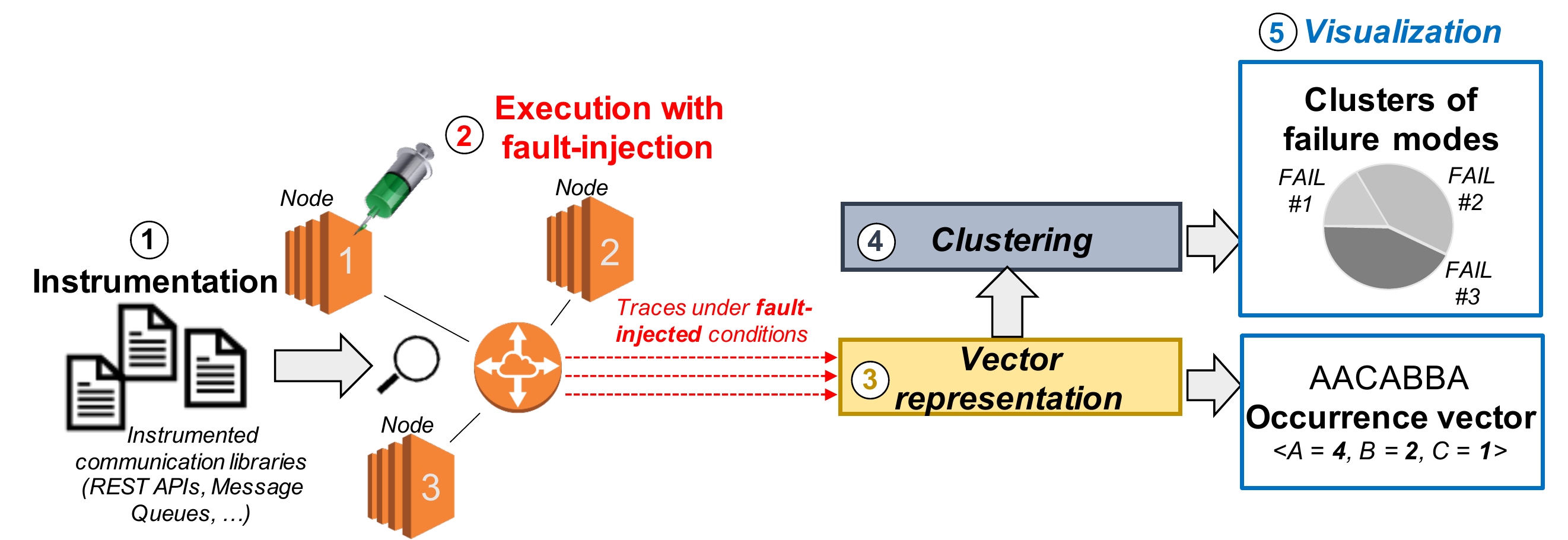}
    \caption{Failure mode analysis based on plain sequences of messages (\textit{SEQ}).}
    \label{fig:SEQ}
\end{figure*}

%\subsection{Problem of Failure Mode Analysis}
\subsection{Overview of cloud computing systems}

A cloud computing system consists of processes distributed across a data center, which cooperate by message passing and remote procedure calls (e.g., through message queues and REST API calls). These systems are quite complex, as they typically consist of software components of millions of lines of code (LoC), which run across dozens of computing nodes. 

In this work, we consider OpenStack as our case study. OpenStack contains a large set of components, each providing APIs to manage virtual resources, and consists of $\sim 20$ million LoC \cite{stackalytics}.
%Figure~\ref{fig:openstack_services} shows a high-level overview of the OpenStack core services.
The three most important components of OpenStack \cite{denton2015learning,solberg2017openstack} are: (i) the \textbf{Nova} subsystem, which provides services for provisioning instances (VMs) and handling their life cycle; (ii) the \textbf{Cinder} subsystem, which provides services for managing block storage for virtual instances; and (iii) the \textbf{Neutron} subsystem, which provides services for provisioning virtual networks, including resources such as \emph{floating IPs}, \emph{ports} and \emph{subnets} for instances. 
In turn, these subsystems include several components (e.g., the Nova sub-system includes \emph{nova-api}, \emph{nova-compute}, etc.), which interact through message queues internally within OpenStack. The Nova, Cinder, and Neutron subsystems provide external REST API interfaces to cloud users.

\subsection{The Problem of Failure Mode Analysis}

Cloud systems can fail in many different ways, and the effects of failures (\emph{failure modes}) are often not known in advance by the developers. In the most subtle cases, the system may be still available to users, but return wrong data, exhibit poor performance, or corrupt the state of resources, leading to poor quality of service.
To identify failure modes of cloud systems, we frame the problem as a \emph{data analysis} task. The input of the analysis is the (failed) executions of the system. Data about failed executions are obtained from fault injection experiments (i.e., faults are introduced to assess fault-tolerance), and from a deployed system in operation. Every execution is represented as a sequence of events (\emph{trace}) that occurred during the execution. This data amounts to thousands of executions and thousands of events in each execution. Therefore, we analyze the traces using \emph{unsupervised machine learning} techniques, to automatically discover recurring patterns among the failures. The effect of failures is valuable for developers, as they can introduce failure management strategies against them.

In our context, an event in a trace represents a \emph{message} exchanged between nodes in the system. In cloud computing systems, the nodes perform or serve a request after receiving messages to provide a service to another node (e.g., through remote procedure calls), and reply with messages to provide the response and results; moreover, nodes use messages to asynchronously notify a new state to other nodes in the system. Therefore, the messages are considered the main observation point for debugging and verification of distributed systems since they reflect the state and the activity of the system \cite{leesatapornwongsa2016taxdc}. 
These messages can be recorded in execution traces for later analysis, using distributed tracing technologies \cite{36356,nedelkoski2019anomaly}, which wrap \emph{communication APIs} invoked by the processes.

To identify failure modes, we perform \emph{clustering} to group the experiments into classes (clusters), where each class represents a distinct failure mode of the system under test. 
In general, clustering algorithms reveal hidden structures in a given data set, by grouping ``similar'' data objects together while keeping ``dissimilar'' data objects in separated classes \cite{xu2005survey}.
Formally speaking, consider a set of $n$ distinct data objects $\{x_1, \ldots ,x_n\}$ and a number of $k$ clusters. A (hard) clustering technique assigns to each data object a label $l_i$ representing its class, with $i \in [1,k]$ \cite{jain1999data}. 
In the context of failure data, a data object represents an execution of the system while it was experiencing a fault. The $i$-th execution is represented by a vector of features $x_i = [ f_1, \ldots, f_d ]$. Each feature is a number that represents how many events of a given type occurred during the execution, with $d$ unique types of events. The number of features easily bump up, due to a large amount of failure data (e.g., hundreds of message types, GBs of log files, thousand of traces, and experiments).

For example, let us suppose that we collected three different message types, $A, B, C$. Let be $x_i = [4,2,1]$ the vector associated to a trace collected during the $i^{th}$ fault injection experiment. This implies that the events $A,B,C$ were observed $4$, $2$ and $1$ times, respectively, during the $i^{th}$ experiment.

The steps for failure mode analysis are summarized in \figurename{}~\ref{fig:SEQ}. We label this basic approach as \textit{SEQ} since it is based on plain sequences of events from fault-injection experiments.

\subsection{Machine Learning Techniques for Failure Mode Analysis}
\label{subsec:background_ML}

\begin{figure*}[t]
\centering
    \includegraphics[width=1.75\columnwidth]{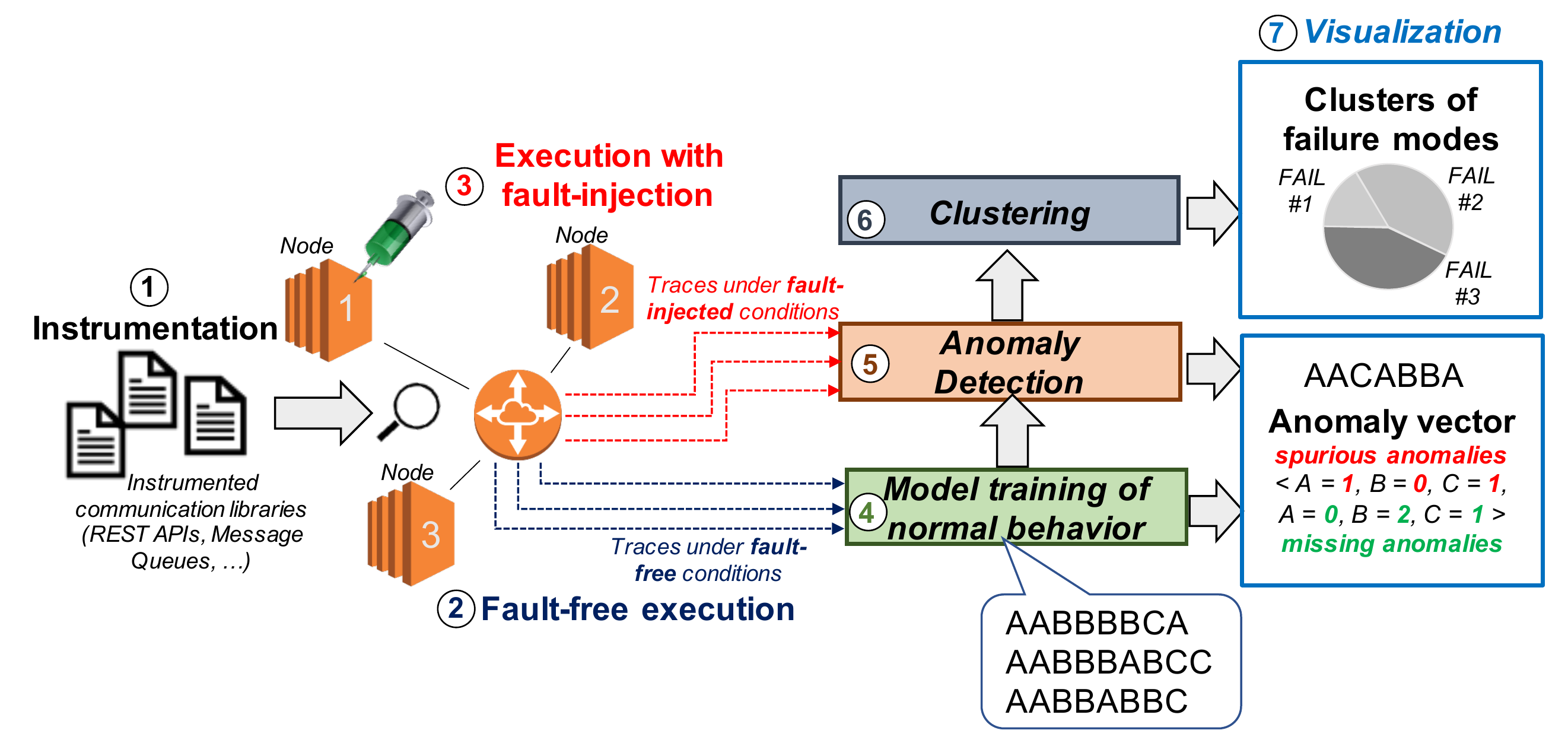}
    \caption{Failure mode analysis based on anomaly detection, using sequence matching and variable-order Markov models (\textit{LCS with VMM}).}
    \label{fig:VMM}
\end{figure*}

Clustering execution traces from cloud systems is a challenging issue, due to the \emph{non-deterministic} behavior of these systems. Even if the system is executed several times in the same way, under the same workload, the timing and the order of messages can unpredictably change, due to delays in communication and computation. 
Thus, there is a need to discriminate between variations in the traces due to different failure modes (which should be divided into different clusters), and ``benign'' variations caused by non-determinism (which should still be grouped in the same cluster).

To improve failure mode analysis, in previous work we proposed an approach based on \emph{anomaly detection} (\figurename~\ref{fig:VMM}), which screens out benign variations from the traces \cite{cotroneo2019enhancing, cotroneo2020fault}. 
The anomaly detection identifies specific events of interest (i.e., failure symptoms) from the large set of events that are generated from fault injection experiments. As a matter of fact, only a few events are actually failure symptoms, such as messages out-of-order or missing, and messages that deliver exceptions. Therefore, it is useful to identify these symptoms and focus clustering on them.

The anomaly detection approach first executes the system several times, using a fault-free workload in which no fault is injected. Thus, any variation in these traces is a benign one, and they are denoted as \textbf{\emph{fault-free traces}}. 
The fault-free traces will be used as a reference for ``normal'' (i.e., non-failure) behavior. 
The approach then performs fault injection experiments and collects \textbf{\emph{fault-injected traces}} from these executions. The approach uses string analysis techniques (\emph{longest common subsequence}, LCS \cite{bergroth2000survey}) to identify common events among fault-free traces and to mark them as non-anomalous.
Then, it applies a \emph{Variable-order Markov Model} (VMM) to analyze non-deterministic variations and to generate a new vector of features (\emph{anomaly vector}). For each message type $t$, the vector includes two features, respectively:

\begin{itemize}
    \item \textbf{Spurious anomalies}: the number of times that an anomalous message of type $t$ appears in the fault-injected trace;
    
    \item \textbf{Omission anomalies}: the number of times that the message type $t$ \emph{does not} appear in the fault-injected trace, but the message should have been occurred according to the probabilistic model. 
\end{itemize}

In total, the number of features is twice the number $d$ of unique event types.  
For example, let us suppose that we observe three different types of messages, $A, B, C$. Let be $x_i = [1,0,1,0,2,1]$ the vector that represents the $i^{th}$ fault-injected trace. These features can be interpreted as follows:
\begin{itemize}
    \item The first three features (valued $1,0,1$) are spurious anomalies. Anomaly detection identified two spurious events, one for the event type $A$ and one for the event type $C$.
    \item The last three features (valued $0,2,1$) are omitted anomalies. Anomaly detection identified three omitted events, two for the event type $B$, and one for the event type $C$.
\end{itemize}

\subsection{Open Issues}

\begin{figure*}[ht]
\centering
    \includegraphics[width=2\columnwidth]{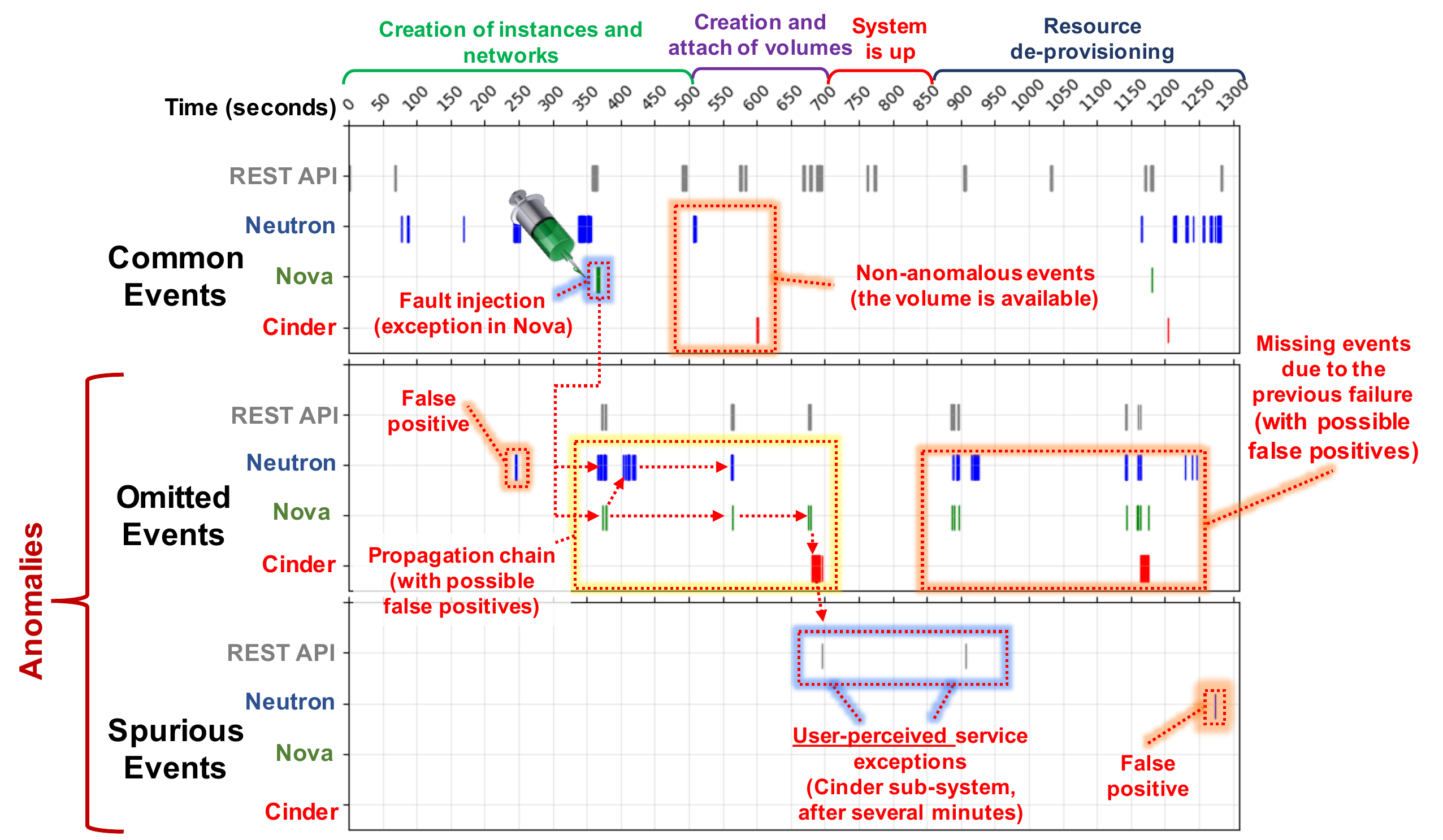}
    \caption{Graphical representation of a fault-injection experiment in the OpenStack cloud computing system.}
    \label{fig:experiment}
\end{figure*}

The previous techniques leverage machine learning to support human analysts at identifying failure modes. From thousands of fault-injection experiments and events, the techniques identify the recurring failure modes (e.g., a dozen of clusters in our previous experience), on which the analyst can focus failure mitigation strategies.

Unfortunately, these techniques require careful tuning by the human analyst to achieve high accuracy. In our previous analysis, we found that accuracy improves when weights are fine-tuned for the most important features.
For example, features representing asynchronous (i.e., non-blocking) messages are more prone to be false positives and less representative of the failure modes; thus, giving a higher weight to features representing synchronous messages (i.e., blocking the caller) increase the accuracy of clustering. Similarly, spurious anomalies on REST API calls often denote exceptions raised by the system, and are more representative of the failure modes.

To better understand this problem, \figurename{}~\ref{fig:experiment} shows a graphical representation of the activation of components in the OpenStack cloud computing system during a fault-injection experiment \cite{cotroneo2019failviz}. The figure divides events into \textit{common}, \textit{spurious}, and \textit{omitted} events, as described in the \S~\ref{subsec:background_ML}. 
%To ease the inspection of the events, the visualization further divides events generated by clients through REST API calls, and events generated internally in the system. 
For simplicity, the figure does not include events for resource monitoring, garbage collection, etc. 
Nevertheless, the figure shows that human analysts must deal with hundreds of events. Some of the events are relevant symptoms of the failure mode, such as exceptions received by the client from REST API calls. Other events are not a symptom of the failure but are benign variations caused by asynchronous updates from Neutron. In order to accurately cluster this failure mode, the features representing REST API calls should be assigned a larger weight than some of the Neutron events, which are non-deterministic and are prone to noise.
%Moreover, the non-determinism of events, which is intrinsic in the nature of the distributed systems, makes the distinction between real anomalies and false alarms even more difficult. 

%Another aspect to consider is a large number of failures to analyze. Since large software systems can be injected with thousands of different faults, fault injection experiments generate thousands of execution traces such as the one of \figurename{}~\ref{fig:experiment}. It is clearly not feasible for the human analyst to repeat the analysis of each fault-injection experiment.

The fine-tuning of weights requires considerable effort by the human analyst, which represents a significant cost and limits the usefulness of the failure mode analysis. Moreover, the tuning requires detailed knowledge of the internals of the system under test, which may be not available for large projects based on software components from different teams and third parties (e.g., commercial vendors). 
Thus, manual-fine tuning of feature weights is a difficult and time-consuming task, and the human analyst needs a different approach for failure mode analysis.

\section{Proposed Approach}
\label{sec:approach}
%ttps://www.dlology.com/blog/how-to-do-unsupervised-clustering-with-keras/

To overcome the open issues of existing techniques, in this work we provide a novel solution to perform failure mode analysis, which does not require a manual effort by the human analyst for feature engineering. To this purpose, we use \emph{deep learning} techniques for generating the features.

\begin{figure*}
    \centering
    \includegraphics[width=1.75\columnwidth]{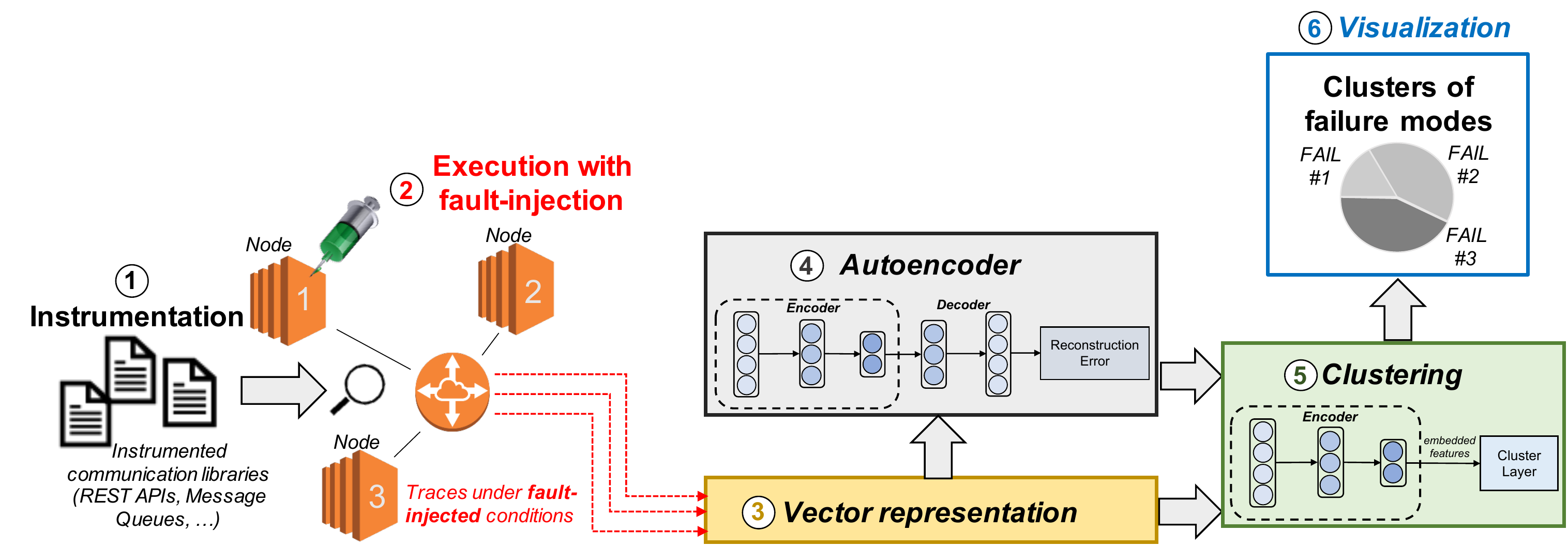}
    \caption{Overview of the proposed solution}
    \label{fig:DEC}
\end{figure*}

Our solution leverages \textit{Deep Embedded Clustering} (DEC), a family of algorithms that performs clustering on the embedded features of an autoencoder \cite{xie2016unsupervised,ghasedi2017deep,guo2017improved,li2018discriminatively,yang2017towards,guo2018deep}. 
The proposed solution (\figurename~\ref{fig:DEC}) uses DEC on the raw vector representations of the fault-injected traces, which are the same ones of the \textit{SEQ} approach discussed before (by replacing the step \circled{4} of \figurename~\ref{fig:SEQ}). This proposed approach relieves the human analyst from fine-tuning the feature weights in the clustering stage, thus saving manual efforts. 

An alternative version of the proposed solution is in combination with anomaly detection, by applying it on anomaly vectors, as in the \textit{LCS with VMM} (by replacing the step \circled{6} of \figurename~\ref{fig:VMM}). In this case, the human analyst invests effort to train an anomaly detection model using fault-free traces, but without manual feature engineering. This combined approach can further improve the accuracy of failure mode analysis. We also analyze this approach in the experimental part of this work.

More in detail, DEC transforms the data with a non-linear mapping $f_\theta : X \rightarrow Z$, where $\theta$ are the learnable parameters, $X$ is the input data (i.e., features about failure), and $Z$ is the embedded feature space (i.e., a new, smaller set of transformed features). We apply a deep neural network (DNN) to parametrize the $f_\theta$ mapping for DEC clusters data by simultaneously learning (i) a set of $k$ clusters centers in the embedded feature space $Z$, and (ii) the parameters $\theta$ of the DNN that performs the mapping between data points (i.e., the input data) and $Z$.
DEC consists of two phases: the initialization of the parameters with a deep autoencoder and the optimization of the parameters. %In the following, we discuss these two phases. 

\subsection{Parameter Initialization}

To initialize the parameters, we use a multi-layer deep autoencoder. An autoencoder is a neural network composed of two parts, an encoder, and a decoder. 
The goal of the encoder is to compress the input features to lower-dimensional features. The decoder part, on the other hand, takes the compressed features as input and reconstructs them as close to the original data as possible. 
Autoencoder is an unsupervised learning algorithm in nature since during training it only uses unlabelled data. 
Our approach applies a fully connected symmetric autoencoder since our vectors are compressed and decompressed in a specular way.

We initialize the autoencoder network layer by layer so that the layers work as a \textit{denoising autoencoder} \cite{vincent2008extracting,lu2013speech} trained to reconstruct the previous layer's output after random corruption of the data.
We set the network input dimension equal to $d$, where $d$ is the number of the vector features (which depends on the number of unique events). 

After the training, we concatenate all the layers of the encoder followed by the layers of the decoder, to form a multi-layer deep autoencoder with a bottleneck coding layer in the middle. 
All layers of the neural network are densely (fully) connected. Our solution is intentionally meant to adopt a typical and regular DNN architecture, to avoid hand-tuning by the human analyst as much as possible. Thus, the value $d$ is the only parameter that depends on specific the failure dataset under analysis.

The autoencoder is trained to minimize the reconstruction loss. Then, we discard the decoder layers, and we apply the encoder layers as our initial mapping between the data space and the feature space. 

To start the clustering phase, we need to initialize the cluster centers. Therefore, we firstly input the initialized DNN with the data points to get embedded data points, and then apply a clustering algorithm in the feature space $Z$ to obtain $k$ initial centroids. Our solution adopts the \textit{K-Medoids}, a clustering method that performs the clustering phase by minimizing the sum of the dissimilarities between objects and a reference point for their cluster. As a reference point, this method uses the \textit{medoid}, i.e., the most centrally located object in a cluster. Therefore, this method is considered less sensitive to outliers than the classical \textit{K-Means}, which takes the mean value of the objects in a cluster as a reference point \cite{arora2016analysis, velmurugan2010computational}.

\subsection{Parameter Optimization}

The approach trains the non-linear mapping $f_\theta$ with two joint objectives: the DNN minimizes the reconstruction error; and, it maximizes inter-cluster variance in the embedded feature space. Towards these goals, the approach alternates between (i) computing a ``soft'' assignment between the current cluster centroids and the embedded data samples (i.e., a vector of probabilities that the sample is a member of each cluster); and (ii) updating the mapping $f_\theta$ and the cluster centroids to maximize inter-cluster variance. We repeat the process until meeting a convergence criterion.

To measure the similarity between the embedded data points and the $k$ centroids, we build a custom layer, named \textit{cluster layer}, to convert the input features to cluster label probability. To quantify the similarity between every embedded point and a centroid (i.e., to assign the probability in the soft assignment), we computed the \textit{Student's t-distribution}.
%Similarly to the \textit{t-SNE} algorithm \cite{van2008visualizing},
%We apply the Student's t-distribution to measure the similarity between an embedded point and a centroid. 
%The clustering layer acts similarly to K-medoids for clustering, and the layer's weights represent the cluster centroids, which can be initialized by training K-medoids.

Then, we recompute the clusters iteratively by learning from the current soft assignment. In particular, the clustering model is trained to minimize the distance between the soft assignments and an artificial ``target'' distribution, which is a transformed version of the probabilities in the soft assignment that widens the gap between the probabilities \cite{peng2019deep}. In our case, we compute the target distribution by raising the soft assignments to the second power and normalizing the values. The approach gives more emphasis on data points assigned with high probability, and at the same time, it also optimizes for the ones with low probability. 
By optimizing for the low distance between the actual soft assignments and the target distribution, we obtain clusters with larger intra-cluster variance, thus improving the cluster quality.

For the optimization, we minimize the \textit{Kullback–Leibler divergence} (KL) between the soft assignments and the target \cite{jabi2019deep}. 
The KL divergence is a loss function that measures the difference between two distributions.
We update the target distributions after a specific number of clustering iterations. The clustering model is then trained to minimize the KL divergence loss between the output of the clustering and the target distribution. %This training can be seen as a self-training strategy \cite{nigam2000analyzing} since, given an initial classifier and an unlabeled dataset, we train the classifier on its high-confidence predictions to label the dataset.
We leveraged the Stochastic Gradient Descent (SGD) with momentum \cite{qian1999momentum} to optimize simultaneously both the cluster centers and the DNN parameters. 
%We jointly optimize the cluster centers and DNN parameters $\theta$ using Stochastic Gradient Descent (SGD) with momentum \cite{qian1999momentum}.
The parameter optimization process stops when a percentage of points below a \textit{convergence threshold} changes the assigned cluster between two iterations in a row. We set the convergence threshold equal to $0.1\%$.

\section{Experiments}
\label{sec:experiments}
In this section, we evaluate the proposed approach in the context of failure data from the OpenStack cloud computing platform. 
We obtain failure data from fault-injection experiments, which were performed on OpenStack version 3.12.1 (release \emph{Pike}), deployed on Intel Xeon servers (E5-2630L v3 @ 1.80GHz) with 16 GB RAM, 150 GB of disk storage, and Linux CentOS v7.0, connected through a Gigabit Ethernet LAN. In particular, in our experiments, we targeted Nova, Cinder, and Neutron subsystems, which are considered the three most important services of OpenStack \cite{denton2015learning,solberg2017openstack}. 
%We injected faults into the three most important services of OpenStack \cite{denton2015learning,solberg2017openstack}: (i) the \textbf{Nova} subsystem, which provides services for provisioning instances (VMs) and handling their life cycle; (ii) the \textbf{Cinder} subsystem, which provides services for managing block storage for virtual instances; and (iii) the \textbf{Neutron} subsystem, which provides services for provisioning virtual networks, including resources such as \emph{floating IPs}, \emph{ports} and \emph{subnets} for instances. 
%In turn, these subsystems include several components (e.g., the Nova sub-system includes \emph{nova-api}, \emph{nova-compute}, etc.), which interact through message queues internally within OpenStack. The Nova, Cinder, and Neutron subsystems provide external REST API interfaces to cloud users.

We injected faults during the execution of OpenStack, by simulating exceptional conditions during the interactions among its components. We targeted the internal APIs used by OpenStack components for managing instances, volumes, networks, and other resources. For example, we injected faults during calls to the \emph{cinder-volume} component within the Cinder subsystem to perform operations on the volumes. 

To define the faults to inject into the target system, we analyzed over $200$ problem reports on the OpenStack bug repository. This analysis allowed us to identify the most recurrent bugs in OpenStack over the last years. In particular, we choose the following faults, which are among the most frequent in OpenStack \cite{cotroneo2019bad}:

%The injected faults represent exceptional cases, e.g., a resource that is not found or unavailable, a processing delay when retrieving a resource, or an incorrect value caused by the user, the configuration, or a bug inside OpenStack. In particular, we considered the following types of faults, which are among the most frequent in OpenStack \cite{cotroneo2019bad}:

\begin{itemize}

    \item \textit{\textbf{Throw exception}}: The target method raises an exception, according to the per-API list of exceptions;
    
    \item \textit{\textbf{Wrong return value}}: The target method returns an incorrect value. In particular, the returned value is corrupted according to its data type (e.g., we replace an object reference with a null reference, or replace an integer value with a negative one);
    
    \item \textit{\textbf{Wrong parameter value}}: The target method is called with an incorrect input parameter. Input parameters are corrupted according to the data type, as for the previous fault type;

    \item \textit{\textbf{Delay}}: The target method returns the result after a long delay. This fault can trigger timeout mechanisms inside the system or can cause a stall.
    
\end{itemize}

\subsection{Workloads}

\begin{table*}[t]
\caption{Workload characteristics}
\label{tab:dictionary_wl}
\small
\centering
\begin{tabular}{>{\centering\arraybackslash}m{1.75cm} >{\centering\arraybackslash}m{1.75cm} >{\centering\arraybackslash}m{2.5cm} >{\centering\arraybackslash}m{2cm} >{\centering\arraybackslash}m{2cm} >{\centering}m{2cm}}
\toprule
\textbf{Workload} & \textbf{Num. unique events} & \textbf{Avg. num. of events per fault-free trace} & \textbf{Num. of total exps.} &  \textbf{Num. of failed exps.} \\ 
\midrule
\textit{DEPL} & 64 & 285 & 1076 & 537\\ 
\textit{NET} & 40 & 252 & 561 & 262\\ 
\textit{STO} & 41 & 109 & 901 &  515\\ 

\bottomrule

\end{tabular}
\end{table*}

We performed three distinct sets of fault injection experiments (\emph{campaigns}), in which we applied three different workloads.

\vspace{1pt}
\noindent
$\rhd$ \textit{\textbf{New deployment workload}} (DEPL): This workload configures a new virtual infrastructure from scratch. It creates VM instances, volumes, key pairs, and a security group; attaches the instances to the volume; creates a virtual network consisting of a subnet and a virtual router; assigns a floating IP to connect the instances to the virtual network; reboots the instances, and then deletes all the created resources. This workload is meant to stress in a balanced way Nova, Cinder, and Neutron subsystems.
        
\vspace{1pt}
\noindent
$\rhd$ \textit{\textbf{Network management workload}} (NET): This workload includes network management operations, to focus interest on the operations related to the virtual networks and, therefore, on the Neutron subsystem. The workload initially creates a network and a VM, then generates network traffic via the public network. After that, it creates a new network with no gateway, brings up a new network interface within the instance, and generates traffic to check whether the interface is reachable. Finally, it performs a router rescheduling, by removing and adding a virtual router resource.
    
\vspace{1pt}
\noindent
$\rhd$ \textit{\textbf{Storage management workload}} (STO): This workload mainly performs the operations related to the storage management of instances and volumes to stress the Nova and Cinder subsystems. It creates a new volume from an image, boots an instance, then rebuilds the instance with a new image before cleaning up the resources. 

\vspace{2pt}

All of these workloads invoke the OpenStack APIs provided by the Nova, Cinder, and Neutron subsystems. We implemented the workloads by reusing integration test cases from the \emph{OpenStack Tempest} project \cite{openstack_tempest} since these tests are already designed to trigger several subsystems and components of OpenStack and their virtual resources. %We selected these workloads to point out propagation effects across subsystems that may be caused by faults. 
We selected these workloads to evaluate our approach in different conditions (i.e., networks operations, storage operations, etc.) and to emphasize the propagation of the failure across different subsystems that can be caused by the injected faults.

To understand when the system experiences a failure, our workload generator performs  \textit{assertion checks} on the status of the virtual resources. For example, the workload assesses the connectivity of the VM instances via SSH, or query the OpenStack API to check the status of the instances, volumes, and networks. 
%In-between calls to service APIs, our workload generator performs \emph{assertion checks} on the status of the virtual resources, to reveal failures of the cloud management system. 
%These assertion checks assess the connectivity of the instances through SSH and query the OpenStack API, to ensure that the status of the instances, volumes, and the network is consistent with the expectation of the tests. 
The checks helped us at manually diagnosing the outcome of every experiment, in addition to logs produced by the system. We used this information to build a \emph{ground truth} of the failures during the experiments, i.e., a reference for evaluating the accuracy of the proposed approach (see the next subsection). 
We consider an experiment as failed if at least one API call returns an error (\textbf{API error}) or if there is at least one assertion check failure (\textbf{assertion check failure}). 
Before every experiment, we re-deploy the cloud management system, remove all temporary files and processes, and restore the OpenStack database to its initial state. These actions are needed to avoid any residual effect of the previous experiments that can impact the current one.
% to ensure that the potential failure is only due to the current injected fault. 
%To this end, we re-deploy the cloud management system, remove all temporary files and processes, and restore the OpenStack database to its initial state.

\subsection{Failure dataset}
To inject faults in Nova, Neutron, and Cinder, we performed a full scan of their source code, using an automated fault injection tool \cite{cotroneo2020profipy}, to identify all injectable API calls. We then checked whether the injectable API calls are indeed executed by the workloads.  In the experimental campaign, we performed one fault injection test for each injectable location that is covered by the workloads. 
In total, we performed $2,538$ fault injection tests, and we observed failures in $1,314$ tests ($52\%$). 
In the remaining tests (33\%), there were neither API errors nor assertion failures, since the fault did not affect the behavior of the system (e.g., the corrupted state is not used in the rest of the experiment). This is a typical phenomenon that often occurs in fault injection experiments \cite{christmansson1996generation,lanzaro2014empirical}; yet, the experiments provided us a large and diverse set of failures for our analysis.

Table~\ref{tab:dictionary_wl} shows, for each workload, the number of event types $d$ observed in the distributed system during the execution of the workloads, the average length of the fault-free sequences (in term of the number of events in the trace), the total number of fault injection experiments for the workload, and the number of experiments that experienced at least one failure.
The number of event types and the total number of events reflects the extent and diversity of the work put on the system. We notice that DEPL is the most extensive workload in terms of both distinct operations and the total number of operations, followed by NET and by STO. 
%These differences among the workloads are meant to evaluate the approach under different levels of complexity and non-determinism.

To every experiment of the fault-injection campaigns, we assigned a label expressing the \textit{failure class}, or \textit{failure mode}, i.e., the type of failure that the system experienced during the experiment. The classes of failure serve as \textit{ground-truth} for evaluating the results of the clustering. A good clustering solution, indeed, should be close to the classification of the ground truth.
Having a reliable ground truth is a common problem in the research problems involving the analysis of the logs. System logs are usually good indicators of system state as they contain reports of events that occur on the several interrelated components of complex systems \cite{lim2008log}. Previous works leveraged the collection of system logs as sources of data, which could be analyzed by a system to make it aware of its internal state
\cite{vaarandi2004breadth,aharon2009one,fu2009execution,makanju2011system}.
Therefore, to assign a failure-class to every experiment, we leveraged the assertion checks and the API errors raised by OpenStack. Furthermore, we investigated the logs of the systems and the anomalies in the traces. To reduce the possibility of errors in manual labeling, all the authors discussed cases of discrepancy, obtaining a consensus on the failure modes.

%We built a ground-truth for the evaluation, by manually labeling the failures. 
%The problem of having a ground truth is a quite common open problem in all the research work dealing with log analysis.
%Data labeled by real system administrators represent the ideal case with the actual ground truth, but this option requires a significant resource commitment from a company.
%Therefore, we mitigated this problem by using the same data source that would be used by a system administrator for analyzing failures, e.g., by OpenStack logs, API Errors experienced by clients, assertion checks from OpenStack developers, anomalies in the traces, etc., to classify the experiments concerning their failure modes, based on our previous experience with OpenStack \cite{cotroneo2019bad}. 
%System logs are usually good indicators of system state as they contain reports of events that occur on the several interrelated components of complex systems \cite{lim2008log}. Previous works leveraged the collection of system logs as sources of data, which could be analyzed by a system to make it aware of its internal state \cite{vaarandi2004breadth,aharon2009one,fu2009execution,makanju2011system}.
%Moreover, to reduce the possibility of errors in manual labeling, multiple authors discussed cases of discrepancy, obtaining a consensus on the failure modes.

\begin{table}[t]
\caption{Failure Mode Classes per Workload}
\centering
%%Table 1
\small
\begin{tabular}{cccc}
\toprule
\textbf{Failure Mode} & \textbf{DEPL} & \textbf{NET} & \textbf{STO} \\
\midrule
\textit{Instance Failure} & 224 & 56 & 320\\ 
\textit{Volume Failure} & 151 & - & 38\\ 
\textit{Network Failure} & 52 & 30 & -\\
\textit{SSH Failure} & 41 & 176 & -\\ 
\textit{Cleanup Failure} & 69 & - & 157\\ 
\textit{No Failure} & 539 & 299 & 386\\ 
\bottomrule
\end{tabular}
\label{tab:ground_truth}
\end{table}

In our experiments, we found the following types of failure modes:

\begin{itemize}
    \item \textit{\textbf{Instance Failure}}: Failure of the operations related to the VM instance. For example, the creation of the virtual machine fails, or the virtual machine is created but it is in not a valid state.
    \item \textit{\textbf{Volume Failure}}: Failure of the operations related to the volume, such as the creation and/or the attach of the volume to the virtual machine, or also the volume is created but it is in an error state.
    \item \textit{\textbf{Network Failure}}: Failure of the operations related to the networks, such as the creation of networks and sub-networks, the association of the IP address to the virtual machine, etc.
    \item \textit{\textbf{SSH Failure}}: Failure to reach the virtual machine via SSH. For example, even if the virtual machine is correctly created and up, it is not reachable for the connection. 
    \item \textit{\textbf{Cleanup Failure}}: Failure related to the operations performed in the last phase of the workload, when the system is not able to serve the requests of deletion of the resources previously created.
    \item \textit{\textbf{No Failure}}: The system can perform all requests without raising any failure during the experiment. 
\end{itemize}

%For all the workloads, we assigned every experiment to a failure class, as shown in \tablename{}~\ref{tab:ground_truth}.  This failure class assignment serves as a reference - or ground truth - for evaluating the results of the clustering. A good clustering solution should be close to the classification of our ground truth. 
Even if we use the same labels for the failure modes across the three workloads, each failure mode should be considered different for each workload since they involve different resources and APIs during execution (e.g., DEPL and STO have both cleanup failures, but with different behaviors).
We found $6$ different failure modes for DEPL and $4$ failure modes for both NET and STO. Since DEPL is our most stressful workload, it is unsurprising to identify a higher number of failure classes among the experiments of this fault-injection campaign. 

We shared the failure dataset on GitHub\footnote{\url{https://github.com/dessertlab/Failure-Dataset-OpenStack}} to help the research community in the application and evaluation of new solutions for clustering the failure modes of the systems. For every experiment of the three fault-injection campaigns, the dataset contains the events exchanged in the system and the corresponding failure label. We shared the representations of experiments with and without the anomaly detection phase (as shown in \figurename{}~\ref{fig:VMM} and \figurename{}~\ref{fig:SEQ}, respectively).

\begin{table*}[t]
\centering
\caption{Purity values of clustering without performing anomaly detection (\textit{SEQ} data). Bolded values are the best performance.}
\small
\label{tab:results_SEQ}

\begin{tabular}{ >{\centering\arraybackslash}m{4cm} >{\centering\arraybackslash}m{2cm} >{\centering\arraybackslash}m{2cm} >{\centering\arraybackslash}m{2cm}}

\toprule

\textbf{Clustering Approach} & \textbf{DEPL} & \textbf{NET} & \textbf{STO}\\
\midrule
\textit{k-medoids w/o fine-tuning}  &  0.70 &  0.80 &  0.80\\
\textit{k-medoids with fine-tuning} &  0.74 & 0.85  & 0.82\\
\textit{DEC} & \textbf{0.86}  & \textbf{0.86} & \textbf{0.92}\\
\bottomrule
\end{tabular}

\end{table*}

\begin{table*}[t]
\centering
\caption{Purity values of clustering on top of anomaly detection (\textit{LCS with VMM)}. Bolded values are the best performance.}
\small
\label{tab:results_VMM}

\begin{tabular}{ >{\centering\arraybackslash}m{4cm} >{\centering\arraybackslash}m{2cm} >{\centering\arraybackslash}m{2cm} >{\centering\arraybackslash}m{2cm}}

\toprule

\textbf{Clustering Approach} & \textbf{DEPL} & \textbf{NET} & \textbf{STO}\\
\midrule
\textit{k-medoids w/o fine-tuning}  &  0.80 &  0.78 &  0.87\\
\textit{k-medoids with fine-tuning} &  \textbf{0.94} &  \textbf{0.86} &  \textbf{0.90}\\
\textit{DEC} & 0.84 & 0.83 & 0.89\\
\bottomrule
\end{tabular}

\end{table*}

\subsection{Evaluation metrics}
\label{subsec:metrics}

To evaluate the quality of the clustering, we compare the cluster assigned to the experiment with the failure class of the experiment defined in our ground truth (\tablename{}~\ref{tab:ground_truth}). 
To associate the clusters to the failure classes, we identify, for every cluster, the failure label with the largest overlap and assign every element in the cluster to the ground-truth class \cite{modha2003feature}.
This evaluation is conservative since it can assign multiple clusters to the same ground truth, but it can not associate the same cluster to different classes of failure.

%We evaluated the clustering by comparing the clusters concerning the failure modes in our ground truth (\tablename{}~\ref{tab:ground_truth}). 
%We compare, for each element in the dataset, the cluster assigned to the element with the actual class of the element, according to the ground truth. We adopt the following rule for the comparison \cite{modha2003feature}: for every cluster generated by the algorithm, we identify the ground-truth class with the largest overlap and assign every element in the cluster to the ground-truth class. In the case of a poor clustering algorithm, multiple clusters may be assigned to the same ground-truth class, but it never assigns the same cluster to multiple ground-truth classes. 

In quantitative terms, let $C$ be the number of ground-truth classes $\{\omega_c\}^C_{c=1}$. 
The \emph{purity} of a cluster is defined as the fraction of elements in the cluster that matches the ground-truth class  \cite{xiong2009k}. Assuming $K$ clusters, for each cluster $\{\tau_k\}^K_{k=1}$ we define $P_k = \nicefrac{1}{n_k} \, \cdot max(n^{c}_k)$, where $n_k$ is the size of the cluster $\tau_k$, and $n^{c}_k$ is the number of elements in the cluster $\tau_k$ that belong to the class with label $w_c$. 
The overall \emph{purity} achieved by a clustering algorithm is the weighted sum of the purities across classes, given by $P = \sum\nolimits_{k=1}^K \nicefrac{n_k}{n} \, \cdot P_k$. Purity ranges between $0$ (total misclassification) and $1$ (perfect clustering).
%The higher the value of purity, the better the clustering quality.

\subsection{Experimental results}

We evaluated our solution in two scenarios:
\begin{itemize}
    \item The deep neural network technique is applied on the raw failure data, without performing any anomaly detection. This is the same data as in the \emph{SEQ} approach (see Section~\ref{sec:background}).

    \item The deep neural network technique is applied on top of anomaly detection, i.e., on the anomaly vectors. This is the same data generated by the \emph{LCS with VMM} approach (see Section~\ref{sec:background}).
\end{itemize}

For each of these cases, we compare the proposed approach (\emph{DEC}) against baselines, in which we apply traditional clustering. For the baselines, we consider both the case of plain features (\emph{k-medoids w/o fine-tuning}), and a manual fine-tuning of the weights of the features (\emph{k-medoids with fine-tuning}), as discussed in Section~\ref{sec:background}. 
We remark that the fine-tuning of the features is a difficult and time-consuming task, due to the exploration of a large number of features (hundreds of event types) and the deeper study of event types in OpenStack (e.g., synchronous and asynchronous events, missing and spurious events, RPC messages and REST APIs, etc.). This exploratory data analysis was performed with Matlab code and took around two weeks of manual effort.

To evaluate different use-cases and conditions,  we applied our solution to perform clustering on the data from the three fault-injection campaigns, one for each workload. 
The input data $X$ is a matrix with the number of rows equal to the number of fault-injection experiments. The columns are dependent on the number of different event types $d$ observed during the execution of the workload. In particular, the number of columns is $d$ when the clustering is applied without the help of the anomaly detection, and $2d$ when the clustering is applied with the anomaly detection (since the algorithm discerns the spurious events from the omitted ones, as explained in \S{}~\ref{subsec:background_ML}.

We set the hyper-parameters to minimize the reconstruction loss.
During the phase of pre-training, we performed a basic tuning of the parameters following the common practices of previous studies \cite{mendoza2019towards,koohzadi2020unsupervised}. We randomly initialized the weights of the layers. The layers were pre-trained for $100,000$ iterations and a drop-out rate set to $20\%$. We trained DEC with additional $100,000$ iterations but without a drop-out rate. 
We set the size of the mini-batch to $256$, the starting learning rate to $10\%$, which is divided by $10$ every $20,000$ iterations, and the weight decay to $0$ \cite{xie2016unsupervised}. 
%In the KL divergence minimization phase, we train with a constant learning rate of $0.01$. 
For each dataset, we tuned the autoencoder by configuring the number and the dimension of the inner layers (between 2 and 4 layers, of decreasing dimension from $d$ to $K$), and the distance metric for clustering ($L1$, city block, and $L2$, euclidean).
Moreover, to initialize the centroids of the clusters, we selected the best solution after running the k-medoids with $30$ repetitions.

\tablename{}~\ref{tab:results_SEQ} and \tablename{}~\ref{tab:results_VMM} show the clustering results, in terms of purity, without and with anomaly detection, respectively. 
The results without anomaly detection (\tablename{}~\ref{tab:results_SEQ}, \textit{SEQ} data) show that the use of the DEC achieves a higher purity compared to traditional clustering, both without and with fine-tuning of feature weights. This behavior applies to each of the three workloads. The scenario without anomaly detection is the most important one since it is the case of the busy system designer that needs quick feedback from fault injection tests, to quickly perform the next iteration of development. For example, the designer may add or revise fault-tolerance mechanisms, and test them again on a new round of fault injection experiments. In these cases, avoiding training an anomaly detection model is useful to speed up data analysis.

In the case of clustering in combination with anomaly detection (\tablename{}~\ref{tab:results_VMM}, data from \textit{LCS with VMM}), the data have already been processed and reduced before clustering. Therefore, clustering achieves better results than using data without anomaly detection. In particular, clustering benefits most in the case of manual fine-tuning of the feature weights, as \textit{k-medoids with fine-tuning} always achieves better results than both the basic \textit{k-medoids w/o fine-tuning} and \textit{DEC}. 
However, these better results come at the cost of manually setting the weights of the features, which requires a deep knowledge of the system internals, and efforts to best tune them concerning the specific workload. 
Instead, the \textit{DEC} approach achieves performance that is close to the case of fine-tuning, with significantly less effort from the human analyst. 
Moreover, \textit{DEC} always returns better results than the basic \textit{k-medoids}, consistently over all the workloads, and both with and without anomaly detection.
Our experiments also pointed out that the standard deviation is below $5\%$, and data are normally distributed around the mean.

To better understand the impact of the clustering on the analysis of failure modes, we inspected the distribution of the failure data samples across the clusters and compared it to the distribution of the actual failure modes (\tablename{}~\ref{tab:ground_truth}). Ideally, the distribution across clusters matches the actual failure modes, so that the human designer can prioritize the development of fault tolerance mechanisms according to the distribution. Moreover, it is sufficient for the human designer to only analyze one or a few experiments from the same class, thus making the analysis more efficient. 
To map the clusters to the failure modes of \tablename{}~\ref{tab:ground_truth}, we followed the approach described in \S{}~\ref{subsec:metrics}. 
We remark that this analysis does not focus on the quality of clusters (i.e., samples misclassified in the wrong cluster), as the previous analysis already provided figures about the purity of the clusters. Here, we focus on the distribution of the clusters that would be presented to the human designer, as the shape of the distribution influences the interpretation of the failure data.

\begin{figure}[t]
    \centering
    \includegraphics[width=1\columnwidth]{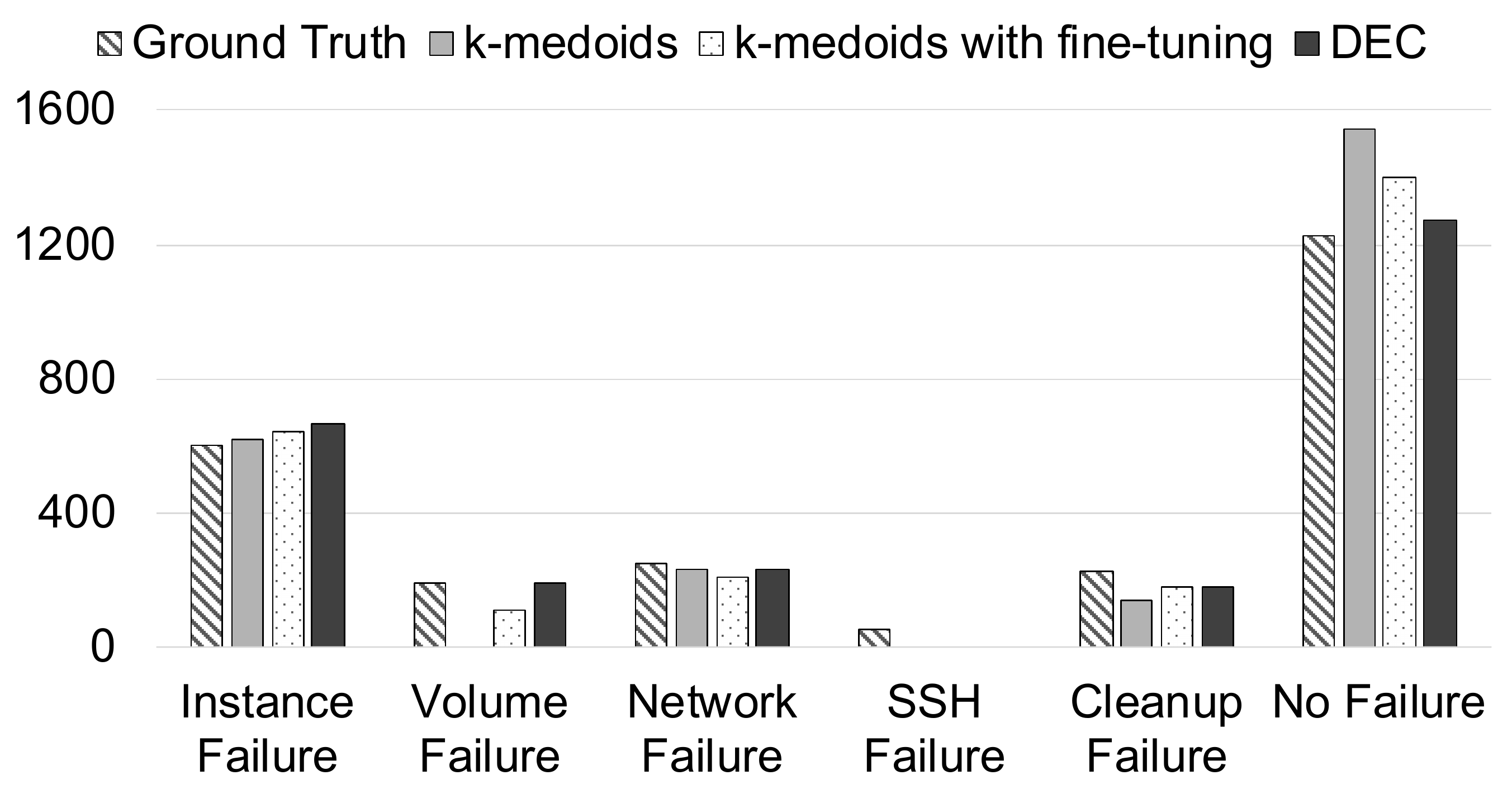}
    \caption{Distribution of failure modes from different clustering techniques (\textit{SEQ} data).}
    \label{fig:failuremodes}
\end{figure}

\figurename{}~\ref{fig:failuremodes} shows the distributions of the clusters for the proposed approach (\emph{DEC}), for the baselines (\emph{k-medoids} with and without fine-tuning), and the actual distribution of the failure modes according to the ground truth. The size of the clusters for \emph{Instance Failure}, \emph{Network Failure}, and \emph{Cleanup Failure} from the clustering techniques are close to the actual frequency of these failure modes. Instead, there are noticeable differences for the remaining failure modes. In the case of \emph{Volume Failure}, the \emph{k-medoids w/o fine-tuning} misses this failure mode, while the cluster from \emph{k-medoids with fine-tuning} is only half of the actual frequency of this failure mode. In the case of \emph{SSH Failure}, which accounts for a minor part of the failures, all of the clustering approaches do not report any failure. We do not attribute this result to the clustering techniques, but to the similarity of events occurring in this failure mode to the ones occurring for \emph{Instance Failure}, which misleads clustering. Instead, we believe that this failure mode could be better analyzed by looking not only at the execution traces but also at additional information sources, such as system logs. Finally, both \emph{k-medoids} with and without fine-tuning over-estimate the cases of \emph{No Failure}, as they report several hundreds of no-failures more than the actual size of this class. This error is the most severe one since it misleads the human designer at believing that the system fails less frequently than the actual truth (e.g., about $-20\%$ of neglected failures). Thus, with the simple \emph{k-medoids}, the analyst would unjustly trust the reliability of the system. Instead, in the proposed approach, the share of cases of \emph{No Failure} is close to the ground truth.

\subsection{Computational Cost}
We evaluated the computational cost of the proposed approach to estimate the overhead introduced by the use of deep learning to cluster the failure data. 
We performed several evaluations, by varying the workloads, the vector representation of the experiments (i.e., with and without the anomaly detection), and the layers of the neural network. 
We found that the use of DEC for clustering introduces an average overhead of $\sim23$ seconds compared to the basic use of the k-medoids. This time includes the initialization of the cluster centers with k-medoids (i.e., the parameter initialization) and the training of the DNN (i.e., the parameter optimization). The standard deviation is high ($\sim75\%$ of the average value) since the configuration of the DNNs impacts the computational cost. 
Nevertheless, the overhead introduced by DEC can be considered acceptable, given that the proposed solution avoids the manual fine-tuning of features, which represents a difficult and time-consuming task.

\section{Conclusion}
\label{sec:conclusion}
In this paper, we presented a novel approach for analyzing failure data from cloud systems, by using unsupervised learning algorithms and deep learning to cluster the failure data into failure classes. 
The proposed approach relieves the human analyst from manually tuning the features to achieve a good performance at clustering failure data. The approach leverages an autoencoder for dimensionality reduction and parameter initialization, in combination with a clustering layer to optimize both the reconstruction error and inter-cluster distance.

We presented results on failure data from the popular OpenStack cloud computing platform. The results show that the proposed approach can achieve performance comparable to, or in some cases even better than, the performance of manually-tuned clustering, which entails a deep knowledge of the domain and a significant human effort. In all cases, the proposed approach performs better than unsupervised clustering when no feature engineering is made on the dataset. 

The approach has been designed to be applied without any a-priori information about the types of features in the failure data, in order to minimize the manual effort. This is especially important when the cloud system is still under active development when multiple versions are updated, tested, and released at a quick pace. However, our approach cannot exceed the accuracy that can be achieved by leveraging the knowledge of the human analyst about the system. Furthermore, since the approach uses deep neural networks, it requires high hardware requirements to keep computational times acceptable, in particular when the amount of data to analyze is very large.

%% The Appendices part is started with the command \appendix;
%% appendix sections are then done as normal sections
%% \appendix

%% \section{}
%% \label{}

\section*{Acknowledgements}
%If you'd like to thank anyone, place your comments here
%and remove the percent signs.
This work has been partially supported by the University of Naples Federico II in the frame of the Programme F.R.A., project id OSTAGE. 
We are grateful to Gabriella Karamanolis for her help in the early stage of this work.

%% If you have bibdatabase file and want bibtex to generate the
%% bibitems, please use
%%
%\bibliographystyle{elsarticle-harv}

\bibliographystyle{elsarticle-num} 
\bibliography{mybibfile}
%\balance
%% else use the following coding to input the bibitems directly in the
%% TeX file.

%\begin{thebibliography}{00}

%% \bibitem[Author(year)]{label}
%% Text of bibliographic item

%\bibitem[ ()]{}

%\end{thebibliography}
\end{document}